%% file: main.tex
\documentclass[10pt,twocolumn,letterpaper]{article}
\usepackage[T1]{fontenc}       
\usepackage{newunicodechar}     
\newunicodechar{–}{--}
\newunicodechar{‐}{-}

\usepackage[pagenumbers]{cvpr}      

\input{preamble}
\input{math_commands.tex}

\usepackage{graphicx}
\usepackage{booktabs}
\usepackage{amsmath}
\usepackage[varbb]{newtxmath}   
\usepackage{multirow}
\usepackage{enumitem}
\usepackage{xcolor}
\usepackage{zref-savepos}

\definecolor{cvprblue}{rgb}{0.21,0.49,0.74}
\usepackage[pagebackref,breaklinks,colorlinks,allcolors=cvprblue]{hyperref}

\def\confName{CVPR}
\def\confYear{2027}

\graphicspath{{figures/}}
\newcommand{\pp}{\,\mathrm{pp}}
\newcommand{\ttt}[1]{\texttt{#1}}
\newcommand{\Vh}{\hat V}
\newcommand{\ph}{\hat\varphi}
\newcommand{\supp}{}

\title{Similar Is Not Useful:\\Selecting In-Context Demonstrations by Closed-Loop Utility}

\author{Jiachen Li\textsuperscript{1}\quad Zhigen Zhao\textsuperscript{2}\quad Shihao Li\textsuperscript{1}\\
Dongmei Chen\textsuperscript{1}\quad Long Chen\textsuperscript{3}\quad Zhuo Xu\textsuperscript{3}\\[4pt]
\textsuperscript{1}University of Texas at Austin\quad \textsuperscript{2}Georgia Institute of Technology\\
\textsuperscript{3}General Intelligent Machines, Inc}

\begin{document}
\maketitle

\begin{strip}
\centering
\includegraphics[width=\linewidth]{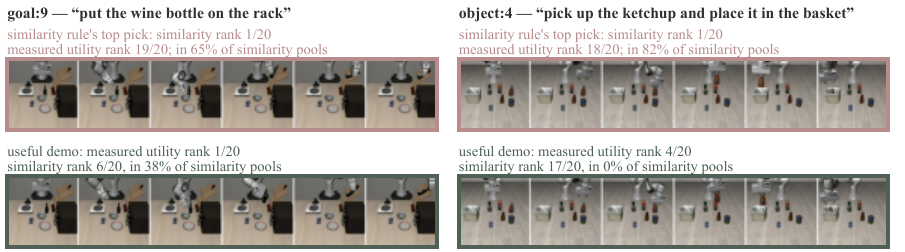}
\captionsetup{hypcap=false}
\captionof{figure}{\textbf{The retriever's favourite demonstration is not the useful one.} Filmstrips of demonstrations from the policy's own pools on two tasks. Rose (top): the demonstration the similarity rule (\ttt{dino}, \cref{sec:setting}) ranks most similar to the initial observations of the test scenes, with the fraction of scenes in which it was placed in the similarity-selected pool and its measured utility rank among the task's 20 demonstrations. Green (bottom): a top-five utility-ranked demonstration of the same task. Utility ranks come from 600 rollouts per task on initial states disjoint from the test scenes (\cref{sec:setting}); single-demonstration ranks are noisy (\cref{sec:why}), and the evidence is the aggregate test in \cref{sec:main}.}
\label{fig:demos}
\captionsetup{hypcap=true}
\end{strip}

\begin{abstract}
Retrieval-augmented vision-language-action (VLA) policies fill their context with the demonstrations whose frames look most like the current observation under frozen DINOv2 embeddings. We ask whether visual similarity selects the demonstrations that make the policy succeed. For a frozen retrieval-augmented VLA (RICL on $\pi_0$-FAST, LIBERO), we make the demonstration pool swappable at run time and score each demonstration's utility by regressing the closed-loop success of 60 fixed-size pools on their membership. On initial states disjoint from those used for scoring, pools of the five highest-utility demonstrations succeed on 66.4\% of 760 paired episodes across 19 tasks, pools of the five most similar on 55.7\%, and random pools on 55.5\%. Utility beats similarity by 10.8 points ($p=3.2\times10^{-6}$; positive on 15 of 18 untied tasks), including on 12 tasks the policy was never primed on, and similarity-selected pools do no better than random ($+0.1$ points). Within a task, a pool's visual similarity to the test scenes tracks its success at Spearman 0.10; across tasks, similarity separates tasks whose scenes differ and falls to chance when two tasks share a scene. The gain comes from on-task closed-loop evidence: best-of-ten direct search over pools recovers two-thirds of it with 100 rollouts. Against the shipped 20-demonstration pool, utility-curated pools of five lead by 7.0 points over 560 paired episodes from two evaluation runs ($p=0.006$), and utility-ordered retrieval inside the full pool leads by 4.5 points ($p=0.055$).
\end{abstract}

\section{Introduction}
\label{sec:intro}

In-context imitation policies place demonstrations in the model's context at inference time and adapt to a task without a gradient step~\cite{sridhar2025ricl,vosylius2025instant,fu2024context,di2024keypoint}. In retrieval-augmented systems such as RICL, a retriever decides which demonstrations enter the context by ranking candidates on visual similarity to the current observation, the distance between frozen vision-encoder embeddings~\cite{oquab2023dinov2}. The prior is sensible: a demonstration that looks like the present scene is likely to be about the present task. Whether the retrieved demonstrations are the ones that make the policy succeed is a closed-loop question, and it is the question this paper answers.

We take a published in-context vision-language-action (VLA) policy, a third-party LIBERO port of RICL on $\pi_0$-FAST~\cite{sridhar2025ricl,pertsch2025fast,liu2023libero}, and hold it frozen. The policy indexes every frame of every demonstration in its pool with DINOv2 features and, at each control step, retrieves the four nearest frames as context. We add one control message to the policy server that swaps the demonstration pool without reloading weights. With that, the expected success of the policy under any pool $S$, $V(S)$, can be estimated by rolling out. Sampling pools of fixed size and regressing the estimated success on membership gives each demonstration a relative utility score $\ph_i$. The question is then whether choosing the pool by $\ph$ beats choosing it by similarity. \Cref{fig:demos} illustrates how the two rankings diverge: on one task the demonstration the similarity rule ranks first is the second-worst of twenty by measured utility; on another, its top choice is in the bottom five while a demonstration it never selects is in the top five.

The answer is that similarity identifies the task and not the useful demonstration within it. Across 19 tasks, pools chosen by measured utility beat pools chosen by similarity by about 11 points, and similarity-chosen pools do no better than random ones. Pools are fixed at five demonstrations, isolating which demonstrations enter the context, and every condition keeps the policy's per-step similarity retrieval inside its pool; the shipped configuration searches all 20 demonstrations. Against that pool, the five highest-utility demonstrations lead by 7.0 points over 560 paired episodes ($p=0.006$), and ordering retrieval inside the full pool by utility leads by 4.5 points ($p=0.055$).

\paragraph{Contributions.}
\begin{enumerate}[leftmargin=1.4em,itemsep=1pt,topsep=2pt]
\item \textbf{An evaluation protocol} for a frozen in-context VLA: a live pool swap that makes $V(S)$ estimable, a fixed-cardinality coalition regression for per-demonstration utility, and a prespecified held-out design that separates the initial states used to fit utility from those used to test selection (\cref{sec:setting}).
\item \textbf{A selection effect that replicates on unseen tasks.} At pool size five, utility-selected pools beat similarity-selected pools by $10.8\pp$ over 19 tasks and 760 paired episodes per cell, with no detectable heterogeneity across tasks; on 12 tasks the policy was never primed on the effect is $+10.2\pp$, against $+11.8\pp$ on the original seven. Similarity-selected pools do no better than random ones (\cref{sec:main}).
\item \textbf{Why similarity fails.} Within a task, a pool's similarity to the test scenes barely tracks its success (mean Spearman 0.10). Across tasks, similarity separates tasks that differ visually and falls to chance when two tasks share a scene (\cref{sec:why,sec:cliff}).
\item \textbf{What buys the gain.} The gain comes from closed-loop evidence on the task: best-of-ten direct search reaches two-thirds of it at 100 rollouts, while removing the worst demonstrations alone did not detectably raise success over random. The effect holds at pool sizes three and five and vanishes at ten (\cref{sec:k,sec:prune,sec:cost}).
\end{enumerate}

\section{Related work}
\label{sec:related}

\paragraph{In-context and retrieval-augmented robot policies.}
Instant Policy~\cite{vosylius2025instant}, ICRT~\cite{fu2024context}, Keypoint Action Tokens~\cite{di2024keypoint}, prompted decision transformers~\cite{xu2022prompting} and RoboPrompt~\cite{yin2025context} condition a policy on demonstrations at inference time. REGENT~\cite{sridhar2025regent} trains a policy on queries interleaved with their retrieved nearest neighbours and adapts to unseen environments in context; RICL~\cite{sridhar2025ricl} brings retrieval-augmented in-context learning to a pretrained VLA, $\pi_0$-FAST~\cite{pertsch2025fast,black2024pi_0}, by post-training on an input in which the $k$ nearest demonstration frames (DINOv2 distance~\cite{oquab2023dinov2}, FAISS index~\cite{douze2025faiss,johnson2019billion}) precede the query; RICL reports physical-robot results on the DROID platform; we study a third-party LIBERO implementation of it and vary only the pool its retriever draws from. Offline retrieval that assembles a fine-tuning set~\cite{du2023behavior,memmel2025strap,lin2024flowretrieval} also ranks by embedding-space similarity, evaluating retrieved data through a fine-tuned policy, whereas we select context for a frozen policy.

\paragraph{Selecting in-context examples for vision models.}
Generalist vision models such as Painter~\cite{wang2023images} and SegGPT~\cite{wang2023seggpt} perform a task specified by example image pairs in context. Visual in-context learning is highly sensitive to the choice of examples; nearest-example retrieval with an off-the-shelf encoder and a retriever trained to maximise in-context performance both improve on random selection~\cite{zhang2023makes}, and prompt-SelF names prompt selection as one of two main factors and selects prompts by pixel-level retrieval~\cite{sun2025exploring}. In those image-level tasks, visual similarity improves on random selection; here the context steers a closed-loop policy whose outcome is a rollout, and similarity-based selection does no better than random selection.

\paragraph{Scoring in-context examples by their effect on the model.}
For language models, the choice and order of in-context examples move accuracy by tens of points~\cite{lu2022fantastically,workrethinking,dong2024survey}; retrieving similar examples helps~\cite{liu2022makes} and retrievers trained on the model's own scores help more~\cite{rubin2022learning,ye2023compositional,zhang2022active}. Closest to our estimator are methods that value in-context examples by their measured effect on outcomes: linear datamodels over example membership that stabilise in-context learning~\cite{chang2023data}, influence-based example selection~\cite{nguyen2023context}, and Shapley-style valuation of demonstrations~\cite{xie2025demoshapley}. We carry outcome-based valuation into a stochastic closed-loop robot setting with a fixed pool cardinality, where the outcome is a rollout and the retriever operates inside the selected pool.

\paragraph{Data valuation and curation.}
Shapley-style valuation~\cite{ghorbani2019data,jia2019towards,kwon2021beta,wang2023data}, datamodels~\cite{ilyas2022datamodels} and gradient-based estimators~\cite{koh2017understanding,pruthi2020estimating,park2023trak} value training data through retraining or its approximation; ATHENA~\cite{xu2026athena} scales influence functions to curating the training data of multitask VLAs. Here the counterfactual is a rollout. Demonstration quality and mixture weights matter for what to train on~\cite{mandlekar2021matters,belkhale2023data,hejna2024re,lin2025data}; we ask what to place in context.

\section{Setting and instrument}
\label{sec:setting}

\begin{figure}[t]
\centering
\includegraphics[width=\linewidth]{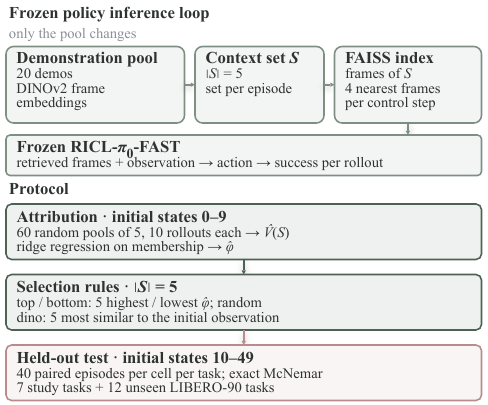}
\caption{\textbf{Protocol.} Top: the frozen policy's inference loop. The demonstration pool feeding the FAISS index is set per episode, while per-step frame retrieval inside the pool is unchanged. Bottom: utility is estimated on initial states 0--9 by regressing the success rate of 60 random size-5 pools (10 rollouts each) on membership; pool-selection rules are then evaluated, all at size 5, on initial states 10--49.}
\label{fig:pipeline}
\end{figure}

\paragraph{Policy.}
The checkpoint is step 10\,000 of a third-party port of RICL to LIBERO published on Hugging Face (\supp\cref{sup:protocol}), a 2.9B-parameter $\pi_0$-FAST model post-trained so that four retrieved demonstration frames, each with its own action, precede the query observation. The retriever loads every demonstration directory in a pool, builds one FAISS index over DINOv2 per-patch embeddings of every frame, and at each control step retrieves the four frames nearest to the current agent-view image. Priming used 12 LIBERO tasks with 20 demonstrations each; we evaluate on both primed and held-out tasks. Each task has a pool of 20 processed demonstrations.

\paragraph{The instrument.}
We added four control messages to the policy server: a pool swap that rebuilds the FAISS index from a chosen set of demonstrations without reloading weights, an embedding endpoint exposing the server's own DINOv2 features (so the similarity baseline uses the retriever's metric), a retrieval-trace counter, and a per-episode RNG reset. Rollouts remain stochastic after the reset (non-deterministic GPU reductions perturb the sampled action tokens), so $V(S)$, the expected success under pool $S$, is estimated by $\Vh(S)$, the success rate over repeated rollouts.

\paragraph{Per-demonstration utility.}
We score demonstrations by sampling 60 pools of fixed size $|S|=5$ from the 20 demonstrations, estimating each $\Vh(S)$ on the same 10 initial states (0--9), and fitting a ridge regression of $\Vh$ on centred membership indicators, $\ph=(X_c^\top X_c+\lambda I)^{-1}X_c^\top y_c$ with $\lambda=1$. Fixing pool size separates demonstration content from cardinality, since the retriever takes the four nearest frames from a single pooled index and a larger pool mechanically supplies nearer neighbours. Because pool size is fixed and the indicators are centred, $\ph_i$ is a \emph{relative} score: adding one demonstration displaces another, and the coefficients sum to zero. Selection uses the two ends of the ranking.

\paragraph{Pool-selection rules and test.}
Four pools of five are compared on every task: \ttt{top} (five highest $\ph$), \ttt{bottom} (five lowest), \ttt{dino} (the five demonstrations with the smallest $L_2$ distance, in the retriever's own DINOv2 embedding, between any of their frames and the test scene's initial observation), and \ttt{random} (seeded per task and scene). All four conditions keep the policy's per-step frame retrieval inside the selected pool; they differ only in which five demonstrations the pool contains. The shipped configuration, the full pool of 20 (\ttt{full}), is evaluated as a separate baseline (\cref{sec:prune}). Cells are evaluated on initial states 10--49, disjoint from the 0--9 used to fit $\ph$, with 40 paired episodes per cell per task; paired cells are compared by the exact McNemar test~\cite{mcnemar1947note} and pooled rates carry Wilson intervals~\cite{wilson1927probable}. A task entered if its full-pool success lay in $[0.25, 0.85]$ on a screen of 10 (LIBERO-90) or 20 (standard suites) trials and passed a prespecified screen on the pool estimates (\supp\cref{sup:protocol}), which excluded one task (\ttt{spatial:7}). The 12 LIBERO-90 tasks receive their own utility fits, so they replicate the whole procedure on unprimed tasks.

\section{Experiments}
\label{sec:exp}

\subsection{Context content is necessary, and the eligible band is narrow}
\label{sec:gate}
\begin{figure}[t]
\centering
\includegraphics[width=\linewidth]{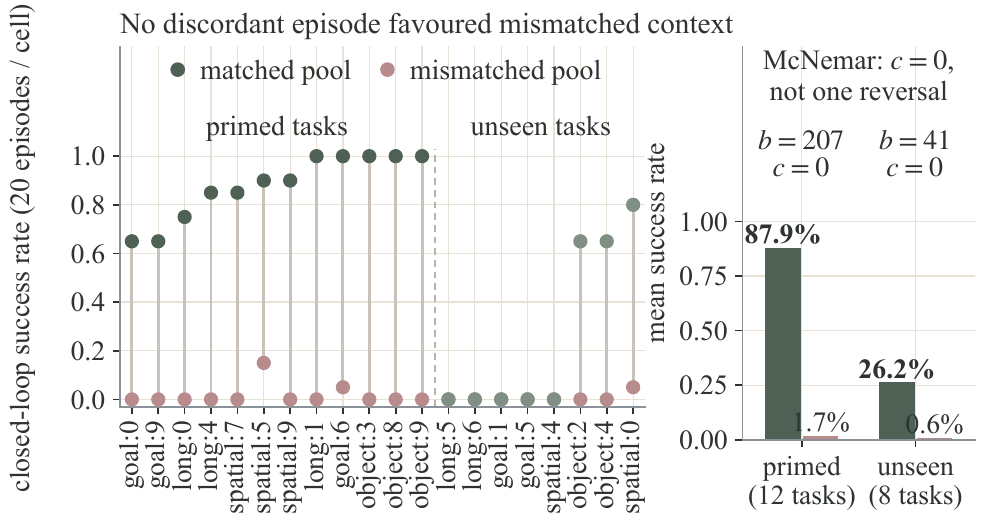}
\caption{Context necessity per task: matched versus mismatched pools, 20 tasks, 20 episodes per cell. No discordant episode favoured mismatched context; held-out tasks fail at the floor and primed tasks saturate.}
\label{fig:gate1}
\end{figure}

We first verify that the policy depends on the demonstrations in its pool. On 8 held-out tasks, swapping a task's own 20 demonstrations for those of a different task (a fixed cyclic shift) moved success from 26.2\% to 0.6\% over 160 paired episodes (matched won 41 discordant pairs, mismatched none, $p<10^{-5}$); on the 12 primed tasks, from 87.9\% to 1.7\% ($b=207$, $c=0$). Primed tasks mostly saturate (7 of 12 above 0.85) and held-out tasks mostly floor (5 of 8 at 0\%); of the 73 remaining LIBERO-90 tasks, 12 (16.4\%) met the screening criterion under a 10-trial full-pool screen, 57 fell below 0.25 and 4 above 0.85 (\cref{fig:gate1}; \supp\cref{fig:ecdf}). Screening yields 7 eligible tasks from the four standard suites (three held-out, four primed) for the attribution study and 12 eligible LIBERO-90 tasks for the replication.

\subsection{Utility beats similarity, and similarity does no better than random}
\label{sec:main}

\begin{table}[t]
\centering
\footnotesize
\setlength{\tabcolsep}{3.2pt}
\caption{Held-out success (\%) at pool size 5 on initial states 10--49, and paired contrasts in percentage points with the exact McNemar $p$ in parentheses. \ttt{top}: five highest measured utility; \ttt{dino}: five most similar to the initial observation; \ttt{bottom}: five lowest utility (anti-selected). 280 episodes per cell for the 7 attribution-study tasks, 480 for the 12 unseen tasks, 760 pooled.}
\label{tab:main}
\begin{tabular}{lccc}
\toprule
& 7 study tasks & 12 unseen tasks & all 19 pooled \\
\midrule
\ttt{top} & \textbf{77.5} & \textbf{60.0} & \textbf{66.4} \\
\ttt{dino} & 65.7 & 49.8 & 55.7 \\
\ttt{random} & 67.5 & 48.5 & 55.5 \\
\ttt{bottom} & 54.6 & 37.5 & 43.8 \\
\midrule
\ttt{top}$-$\ttt{dino} & $+11.8$ (3e$-$3) & $+10.2$ (4e$-$4) & $\mathbf{+10.8}$ (3e$-$6) \\
\ttt{dino}$-$\ttt{random} & $-1.8$ (0.70) & $+1.3$ (0.70) & $+0.1$ (1.0) \\
\ttt{top}$-$\ttt{bottom} & $+22.9$ (1e$-$8) & $+22.5$ (2e$-$15) & $+22.6$ (7e$-$23) \\
\bottomrule
\end{tabular}
\end{table}

\begin{figure}[t]
\centering
\includegraphics[width=\linewidth]{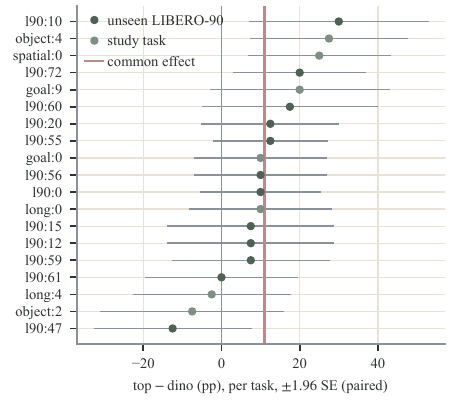}
\caption{\textbf{Per-task \ttt{top}$-$\ttt{dino}, 19 tasks}, with $\pm1.96$ standard errors computed from each task's discordant counts as $\sqrt{b+c}/n$ (a null-based paired standard error; \supp\cref{tab:pertask}). The rose line is the inverse-variance common effect ($+11.0\pp$); Cochran's $Q$ does not reject a common effect ($p=0.31$, $I^2=12\%$).}
\label{fig:main}
\end{figure}

\Cref{tab:main} and \cref{fig:main} give the main result. On the 7 attribution-study tasks the utility-selected pool succeeds on 77.5\% of held-out episodes, the similarity-selected pool on 65.7\%, a random pool on 67.5\% and the anti-selected pool on 54.6\%. Under a first evaluation seed, the same fitted pools gave 77.1 / 69.3 / 68.6 / 54.6 (\ttt{top}$-$\ttt{dino} $+7.8\pp$), with \ttt{top} and \ttt{bottom} within half a point of the second seed reported in \cref{tab:main}. The 12 LIBERO-90 tasks the policy was never primed on provide an independent replication: the contrast is $+10.2\pp$ against $+11.8\pp$, despite lower absolute success. Pooled over 19 tasks, utility selection beats similarity selection by $10.8\pp$ ($p=3.2\times10^{-6}$). Treating the task as the unit gives the same answer: the mean per-task difference is $10.8\pp$ (95\% CI $5.4$ to $16.2$; $t_{18}=4.18$, $p=5.6\times10^{-4}$), and 15 of the 18 untied tasks favour utility (sign test $p=0.008$). Similarity selection does no better than random selection: $+0.1\pp$, 141 discordant pairs won against 140 lost, paired 95\% CI $-4.2$ to $+4.5\pp$.

\ttt{top}$-$\ttt{bottom} ($+22.6\pp$) measures how far demonstration choice can move success; \ttt{top}$-$\ttt{dino} ($+10.8\pp$) measures the gain over similarity at matched pool size. \Cref{sec:prune} compares both rules with the shipped 20-demonstration pool.

\paragraph{Per-task estimates.}
Of the 19 per-task \ttt{top}$-$\ttt{dino} contrasts, 15 are positive, one is zero and three are negative (\ttt{object:2} $-7.5$, \ttt{long:4} $-2.5$, \ttt{libero\_90:47} $-12.5\pp$). An inverse-variance fixed-effect model gives a common effect of $+11.0\pp$ with Cochran's $Q=20.4$ on 18 degrees of freedom ($p=0.31$, $I^2=12\%$)~\cite{cochran1954combination,higgins2002quantifying}: the test does not reject a common effect, and the observed spread of the per-task contrasts (SD $10.9\pp$) is close to their mean standard error ($9.9\pp$) at 40 episodes per task.

\begin{figure*}[t]
\centering
\includegraphics[width=0.86\linewidth]{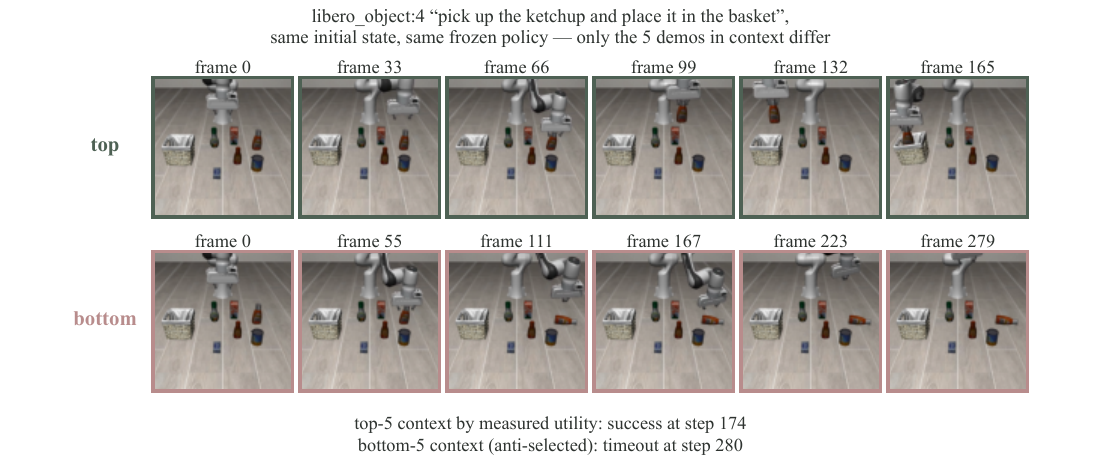}
\caption{\textbf{Same initial state, same frozen policy, different five demonstrations in context} (\ttt{libero\_object:4}): success by step 174 with the five highest-utility demonstrations, a knocked-over bottle and timeout with the five lowest. \Cref{tab:main} gives the statistics over 760 paired episodes per cell.}
\label{fig:rollout}
\end{figure*}

\paragraph{What good context changes.}
Utility-selected pools succeed in 217 of 280 episodes against 153 for anti-selected pools; the two cumulative success curves coincide below about 130 steps and separate after it (\supp\cref{fig:ecdf}), so the gain comes from episodes that finish late. \Cref{fig:rollout} shows one pair of rollouts.

\subsection{Selection pays at small pools and saturates at ten}
\label{sec:k}

\begin{figure}[t]
\centering
\includegraphics[width=\linewidth]{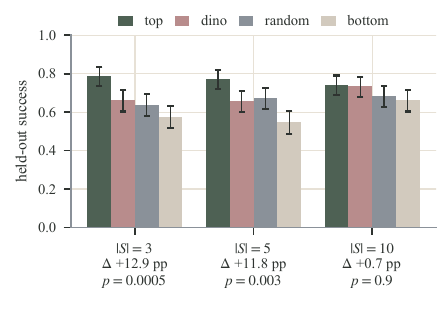}
\caption{\textbf{Context size.} The four cells at pool sizes 3, 5 and 10 on the same 7 tasks, 280 episodes per cell, with Wilson 95\% intervals; $\Delta$ is \ttt{top}$-$\ttt{dino} with its exact McNemar $p$.}
\label{fig:ksweep}
\end{figure}

\Cref{fig:ksweep} repeats the four cells at pool sizes 3, 5 and 10 on the same 7 tasks and 280 episodes per cell (sizes 3 and 10 with rollout seed 7, size 5 with seed 8). \ttt{top}$-$\ttt{dino} is $+12.9\pp$ at size 3 ($p=4.7\times10^{-4}$), $+11.8\pp$ at size 5 and $+0.7\pp$ at size 10 ($p=0.92$). The effect is a boundary, flat from three to five and gone at ten, and the mechanism is visible in the cells: the ten highest and ten lowest of twenty demonstrations share much of the pool's middle, so \ttt{bottom} rises to 66.1\% and \ttt{top} falls to 74.3\% (\ttt{top}$-$\ttt{bottom} $+8.2\pp$, $p=0.027$). \ttt{dino}$-$\ttt{random} stays near zero at every size ($+2.1$, $-1.8$, $+5.0\pp$).

\subsection{Pruning and the shipped 20-demonstration pool}
\label{sec:prune}

\begin{figure}[t]
\centering
\includegraphics[width=\linewidth]{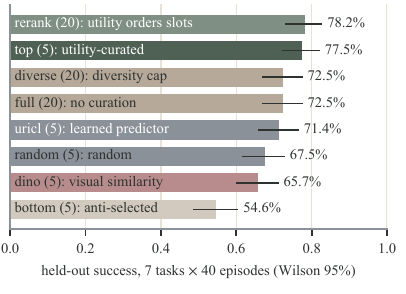}
\caption{\textbf{All strategies} on the 7 attribution-study tasks, 280 episodes each; the pool size is in parentheses. Strategies at pool size 20 use the full pool; \ttt{rerank} orders retrieval slots by $\ph$ fit on the same tasks; \ttt{uricl} is the rollout-free utility predictor of \cref{sec:cost}. One evaluation run per strategy; \cref{sec:prune} adds a second run of \ttt{full}, \ttt{rerank} and \ttt{top}.}
\label{fig:strategies}
\end{figure}

\begin{figure*}[t]
\centering
\includegraphics[width=\linewidth]{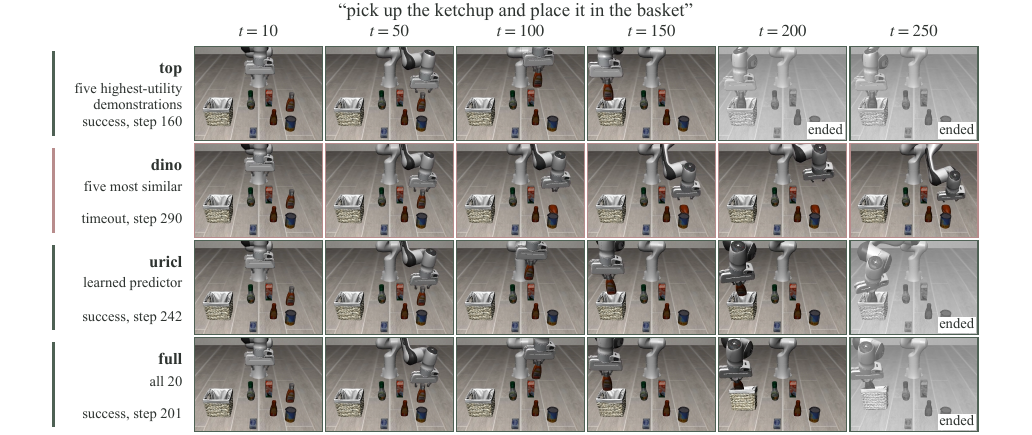}
\caption{\textbf{One initial state, four pools.} Recorded rollouts of the frozen policy on \ttt{libero\_object:4} from the same initial state with the \ttt{top}, \ttt{dino}, \ttt{uricl} and \ttt{full} pools: the similarity-selected pool times out while the other three succeed. This initial state is one of four declared before recording and is shown for illustration; frame times are approximate control steps read from the recorded videos. Success rates over 280 episodes per rule are in \cref{fig:strategies}.}
\label{fig:rolloutcompare}
\end{figure*}

Removing the worst demonstrations did not detectably help on its own. A \ttt{prune} cell that draws five at random from the fifteen demonstrations outside the bottom group scores 70.0\% against 67.9\% for \ttt{random} in the same run ($+2.1\pp$, $b=54$, $c=48$, $p=0.62$; \supp\cref{fig:prune}), whereas \ttt{top} leads \ttt{random} by $10.0\pp$ in \cref{tab:main} ($p=0.008$).

The shipped configuration loads all 20 demonstrations and scores 72.5\% (\cref{fig:rolloutcompare} shows one recorded initial state under four of these rules). Both utility-based rules score above it, the curated \ttt{top} five at 77.5\% ($+5.0\pp$, $p=0.19$) and ordering the retrieval slots by $\ph$ inside the full pool (\ttt{rerank}, $\ph$ fit on the same task's states 0--9) at 78.2\% ($+5.7\pp$, $b=46$, $c=30$, $p=0.085$), the highest of any rule. To increase power beyond 280 paired episodes, we ran a prespecified repeat of \ttt{full}, \ttt{rerank} and \ttt{top} under a new evaluation seed shared by all three: they scored 71.1\%, 74.3\% and 80.0\%. Pooled over both runs (560 pairs per contrast), the curated \ttt{top} five lead the shipped pool by $7.0\pp$ ($b=117$, $c=78$, $p=0.006$), and \ttt{rerank} leads it by $4.5\pp$ ($b=91$, $c=66$, $p=0.055$; \supp\cref{sup:repeat}). Similarity curation moves the other way, to 65.7\% ($-6.8\pp$, $p=0.08$), the lowest rule short of anti-selection, and a slot-diversity cap without utility information matches the full pool exactly (72.5\%). All strategies are in \cref{fig:strategies}.

\subsection{Within a task, similarity barely tracks success}
\begin{figure}[t]
\centering
\includegraphics[width=\linewidth]{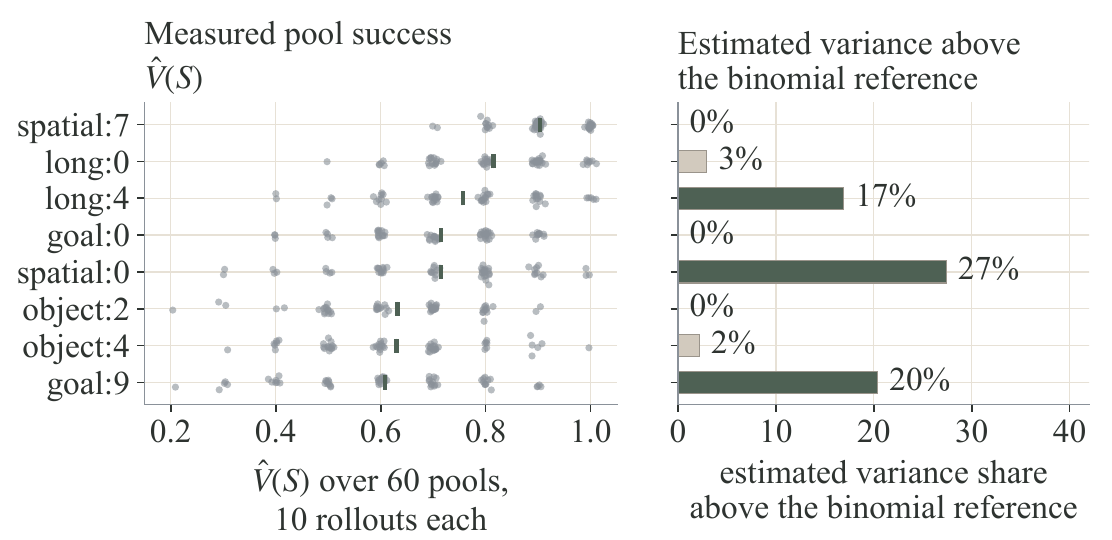}
\caption{The observed variance of $\Vh(S)$ over pools against the binomial reference variance from 10 rollouts per pool; the estimated share above the reference ranges from 0\% to 27\% .}
\label{fig:noise}
\end{figure}
\label{sec:why}

\begin{figure}[t]
\centering
\includegraphics[width=\linewidth]{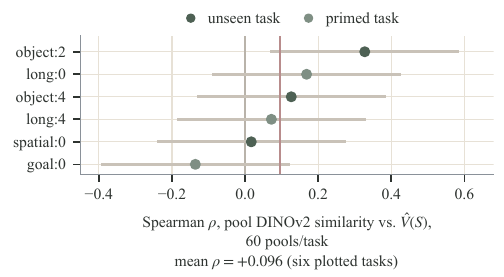}
\caption{\textbf{Pool similarity weakly correlates with measured pool success.} Spearman correlation between a pool's mean DINOv2 similarity to the test scenes and its estimated success $\Vh(S)$, 60 pools per task at fixed size 5 (standard error about 0.13 per task); the six tasks whose per-pool records survive are shown, and the mean over all seven (including \ttt{goal:9}, $\rho=+0.134$) is $+0.10$.}
\label{fig:why}
\end{figure}

Across 60 pools per task at fixed size 5, the Spearman correlation between a pool's mean similarity to the test scenes and its estimated success is $-0.14$, $+0.02$, $+0.07$, $+0.13$, $+0.13$, $+0.17$ and $+0.33$ on the seven tasks, a mean of $+0.10$ (\cref{fig:why}). Only \ttt{object:2} ($\rho=+0.33$) exceeds two standard errors, and it is also one of the three tasks on which similarity selection beat utility selection (\cref{sec:main}). At 10 rollouts per pool, the binomial reference accounts for 73--100\% of the observed variance of $\Vh(S)$ (\cref{fig:noise}), so most of the apparent spread between pools is rollout noise and the demonstration signal lives in the averages that the regression pools across 60 draws.

The held-out separation between \ttt{top} and \ttt{bottom} holds even though individual $\ph$ scores have high sampling variance: recomputing $\ph$ from initial states 0--4 and 5--9 on the same pools gives a mean split-half Spearman of $0.14$ over the 20 attributed tasks (\supp\cref{tab:reliability}), and the two halves' top-five sets overlap by 0.20 (Jaccard). Because $\ph$ is fit on initial states 0--9 and every selection cell is evaluated on 10--49, noise in $\ph$ is independent of the evaluation; a ranking carrying no information would give \ttt{top} and \ttt{bottom} the same expected held-out success, and the $22.6\pp$ gap measures the information the ranking carries.

\subsection{Across tasks, similarity separates scenes, not goals}
\label{sec:cliff}

\begin{figure}[t]
\centering
\includegraphics[width=\linewidth]{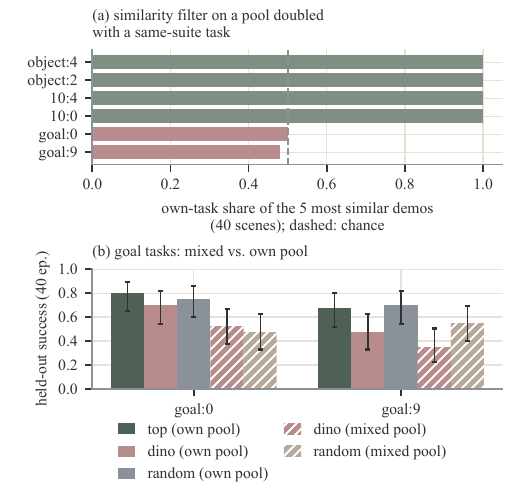}
\caption{\textbf{Similarity identifies the task when scenes differ and not when they coincide.} (a) Each task's pool is doubled with the 20 demonstrations of its same-suite partner among the attributed tasks: three task pairs, each tested in both directions (chance share of own-task demonstrations among the five most similar: 0.5). In the \ttt{object} and \ttt{libero\_10} pairs, the five nearest demonstrations are own-task in 40 of 40 scenes; in the \ttt{goal} pair, whose two tasks share one scene, the share is at chance. No rollouts are needed for this panel. (b) On the two \ttt{goal} tasks, held-out success of similarity and random selection from the mixed pool (hatched) beside the own-pool cells; 40 episodes each.}
\label{fig:cliff}
\end{figure}

Pairing each of the six same-suite tasks with its partner doubles the pool to 40 demonstrations and asks, without rollouts, how many of the five demonstrations nearest under the retriever's metric belong to the task at hand (\cref{fig:cliff}a). The two \ttt{object} tasks and the two \ttt{libero\_10} tasks differ visually, and in both directions of both pairs all five are own-task in all 40 scenes. The two \ttt{libero\_goal} tasks share a scene and differ only in goal, and there the share is at chance (2.50 and 2.40 of 5); on \ttt{goal:9} the single nearest demonstration belongs to the other task in 57.5\% of scenes. The failure is abrupt, a cliff between a perfect filter and a coin flip. On the two \ttt{goal} tasks, similarity selection from the mixed pool scores 52.5\% and 35.0\% against 70.0\% and 47.5\% from the own pool, and random selection from the mixed pool scores 47.5\% and 55.0\% (\cref{fig:cliff}b), so similarity does not recover the task that the mixed pool hides.

\begin{table*}[t]
\centering
\small
\caption{Hypotheses that looked supported mid-study and were withdrawn.}
\label{tab:retracted}
\begin{tabular}{p{0.44\linewidth}p{0.50\linewidth}}
\toprule
Claim & Evidence motivating withdrawal \\
\midrule
$\Vh(S)$ spans 0.30--1.00, so demonstration choice is decisive & the binomial reference accounts for 73--100\% of the observed variance at 10 rollouts per pool \\
Suite structure (goal $<$ spatial $<$ long $<$ object) explains when similarity works & The pre-declared confirmation task \ttt{goal:9} returned $\rho=+0.13$ against \ttt{goal:0}'s $-0.14$; within-suite spread exceeds the between-suite gap \\
The effect is monotone in selection pressure (size $3>5>10$) & Held at 3 tasks per condition, not at 7 \\
$\mathrm{corr}(\rho,\ \ttt{dino}-\ttt{random})$ predicts where similarity helps & Decayed $0.999\to0.925\to0.837$ as tasks were added; sign flips on the size-10 subset \\
$\ph$ from 15 pools matches $\ph$ from 60 & Single draw; at five draws it is 3.4\,pp lower \\
Direct search fails, so the attribution machinery buys the gain & Single draw; repeated-draw mean 73.8\%, versus 65.7\% for similarity and 74.1\% for 15-pool $\ph$ \\
\bottomrule
\end{tabular}
\end{table*}

\begin{figure}[t]
\centering
\includegraphics[width=\linewidth]{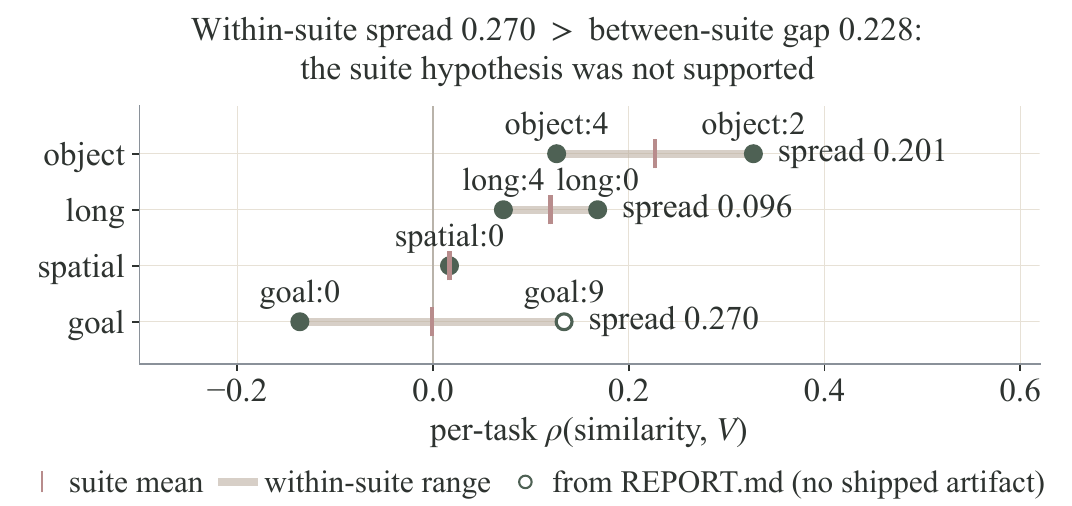}
\caption{The suite-structure hypothesis and the pre-declared confirmation task that overturned it (\cref{tab:retracted}).}
\label{fig:suite}
\end{figure}

\subsection{The gain is bought with on-task rollouts}
\label{sec:cost}

\begin{figure}[t]
\centering
\includegraphics[width=\linewidth]{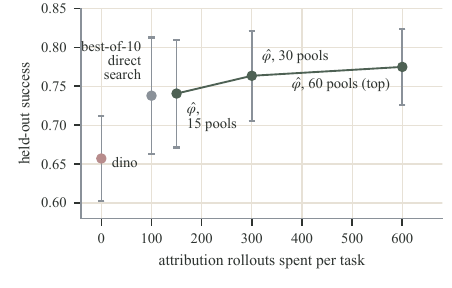}
\caption{\textbf{Held-out success against attribution rollouts spent per task}, descriptive. Grey: best-of-10 direct search. Green: $\ph$ fit from 15, 30 or 60 pools. Points at 100--300 rollouts are means over 5 sub-samples per task (35 task-draw cells per ridge budget; 29 for direct search, whose draws are de-duplicated to 3--5 per task) with the mean within-task SD across draws; endpoints are single cells with Wilson intervals. Sub-samples are compared with the same 280-episode reference cells.}
\label{fig:budget}
\end{figure}

Most of the gain is available for a fraction of the 600 rollouts (about 8.4 GPU-hours) per task that fitting $\ph$ costs, because it comes from closed-loop evidence on the task itself, not from the estimator that summarises it (\cref{fig:budget}). Best-of-ten direct search, which evaluates ten random pools for 10 rollouts each and keeps the best, reaches 73.8\% at 100 rollouts, two-thirds of the way from similarity (65.7\%) to the full fit (77.5\%). Fitting $\ph$ from 15 and 30 pools reaches 74.1\% and 76.4\%, so success rises steadily with the rollout budget, and at a matched budget of 100--150 rollouts direct search and the ridge fit perform alike. Draws at a fixed budget spread by 6--7.5\,pp, so a single draw can mislead in either direction.

A reranker that predicts utility from rollout-free features lands between the two rules. Ridge regression from nine per-demonstration features (query-distance statistics and their rank, frame count, action-magnitude mean and standard deviation), trained leave-one-task-out on 1\,400 rows, reaches a median Spearman of $0.059$ with $\ph$ on the held-out task and selects pools that succeed 71.4\% of the time, $5.7\pp$ above similarity ($p=0.13$) and $6.1\pp$ below utility ($p=0.08$). Its features are functions of similarity and motion statistics, and, consistent with \cref{sec:why}, the information that separates useful demonstrations lies outside them.

\section{Limitations}
\label{sec:limits}

The study covers one checkpoint of one third-party RICL port on LIBERO. Three other published checkpoints of the port fail the context-necessity check on four prespecified tasks (\supp\cref{sup:ckpt2}), so the selection experiment could not be run on them; physical-robot transfer is untested. LIBERO itself fails several benchmark-validity diagnostics in a recent audit~\cite{jiang2026we}; our comparisons are paired contrasts between selection rules for one frozen policy on identical initial states. Against the shipped 20-demonstration pool, the 4.5-point lead of utility-ordered retrieval is not resolved at 560 paired episodes (\cref{sec:prune}). The per-task utility estimates have split-half agreement of 0.14, so they rank demonstrations usefully only in aggregate. One task, \ttt{long:4}, is negative at every pool size because its test split is easier than its fitting split (random cell 90\% on the first evaluation seed against a 76\% attribution-time estimate) and is retained under the prespecified rule. Three per-task episode files from the first evaluation seed are lost, so that seed is reported in aggregate. \Cref{tab:retracted} lists six intermediate readings that repeated measurement overturned, most of them trends inferred from three tasks per condition; \cref{fig:suite} shows one.

\section{Conclusion}
\label{sec:conclusion}

Choosing in-context demonstrations by measured closed-loop utility improved a frozen in-context VLA by about 11 points over choosing them by similarity at matched pool size, across 19 LIBERO tasks and on tasks the policy had never been primed on, while similarity-based choice did no better than random choice. The utility-selected five also beat the policy's shipped 20-demonstration pool, by 7 points. Similarity identifies which task a demonstration belongs to; closed-loop evidence identifies which demonstrations of that task help. A hundred on-task rollouts already buy two-thirds of the gain, so a retriever for in-context robot policies benefits from an outcome signal alongside visual distance.

\paragraph{Reproducibility.}
The campaign scripts, the four server patches and the per-episode records behind every reported cell except the first evaluation seed's are provided as supplementary material (\supp\cref{sup:repro}); the seven-task study used 170 GPU-hours.

\label{endofmain}\zsavepos{endmain}

\section*{LLM Usage Statement}
AI tools, including Claude Code, OpenAI Codex, and Google Gemini, were used to assist with coding, manuscript editing, and formatting. All implementations, results, and conclusions were independently reviewed and verified by the authors. AI tools were not used for data generation, theoretical development, or proofs. The authors take full responsibility for the content.

{
    \small
    \bibliographystyle{ieeenat_fullname}
    \bibliography{references}
}

\clearpage
\appendix
\renewcommand{\thesection}{S\arabic{section}}\renewcommand{\thefigure}{S\arabic{figure}}\renewcommand{\thetable}{S\arabic{table}}\renewcommand{\theequation}{S\arabic{equation}}
\setcounter{section}{0}\setcounter{figure}{0}\setcounter{table}{0}\setcounter{equation}{0}
\begin{center}{\Large\bf Appendix}\end{center}
Section, figure and table numbers with the prefix S refer to this document; all other numbers refer to the main text.

\section{Protocol details}
\label{sup:protocol}

\begin{table*}[t]
\centering
\footnotesize
\caption{Evaluation contract.}
\label{tab:contract}
\begin{tabular}{p{0.2\linewidth}p{0.75\linewidth}}
\toprule
Policy & RICL-$\pi_0$-FAST-LIBERO, checkpoint step 10\,000 of the third-party LIBERO port published on Hugging Face by user \ttt{ryanhoangt} (\ttt{pi0\_fast\_libero\_icl\_priming\_ckpt\_10000}, image-key-fixed normalisation statistics); config \ttt{pi0\_fast\_libero\_ricl}, four retrieved frames \\
Pools & 20 processed demonstrations per task from the port's preprocessed data (DINOv2 per-patch frame embeddings, 49\,152-d) \\
Attribution & 60 pools of size 5 sampled uniformly (seed 0); 10 rollouts each on initial states 0--9; ridge $\lambda=1$ on centred indicators \\
Selection cells & size 5; initial states 10--49; 40 paired episodes per cell; \ttt{random} seeded per (evaluation seed, task, scene); the second evaluation seed reuses the fitted scores and the utility-selected pools and resamples the random pools \\
Similarity & per scene, minimum $L_2$ between any frame of a demonstration and the scene's first-frame embedding; the five nearest demonstrations form \ttt{dino} \\
Step caps & spatial 220, object 280, goal 300, long 520, plus 10 settling steps \\
Screening & full-pool success in $[0.25, 0.85]$ on 10 (LIBERO-90) or 20 (standard suites) screening trials; exclusion if pool-estimate SD $<0.10$ or $>25\%$ of pools at 1.0 \\
Tests & exact McNemar on discordant pairs; Wilson intervals; inverse-variance fixed-effect model with Cochran's $Q$ for heterogeneity \\
\bottomrule
\end{tabular}
\end{table*}

\paragraph{Tasks.} Attribution-study tasks: held-out \ttt{spatial:0}, \ttt{object:2}, \ttt{object:4}; primed \ttt{goal:0}, \ttt{goal:9}, \ttt{libero\_10:0}, \ttt{libero\_10:4}. \ttt{spatial:7} was attributed and excluded by the prespecified rule (pool-estimate SD 0.075; 27\% of pools at 1.0). Unseen LIBERO-90 tasks: indices 0, 10, 12, 15, 20, 47, 55, 56, 59, 60, 61, 72.

\paragraph{Retrieval consistency check.} The four nearest frames from a 16-neighbour search equalled the four from a 4-neighbour search in 80 of 80 queries at both pool sizes on the two tasks tested, so the over-retrieve variants (\ttt{rerank}, \ttt{diverse}) are compared directly with \ttt{full} (\cref{sec:prune}).

\section{Per-task results}
\label{sup:pertask}

\begin{figure*}[t]
\centering
\begin{minipage}[t]{0.49\linewidth}\centering\includegraphics[width=\linewidth]{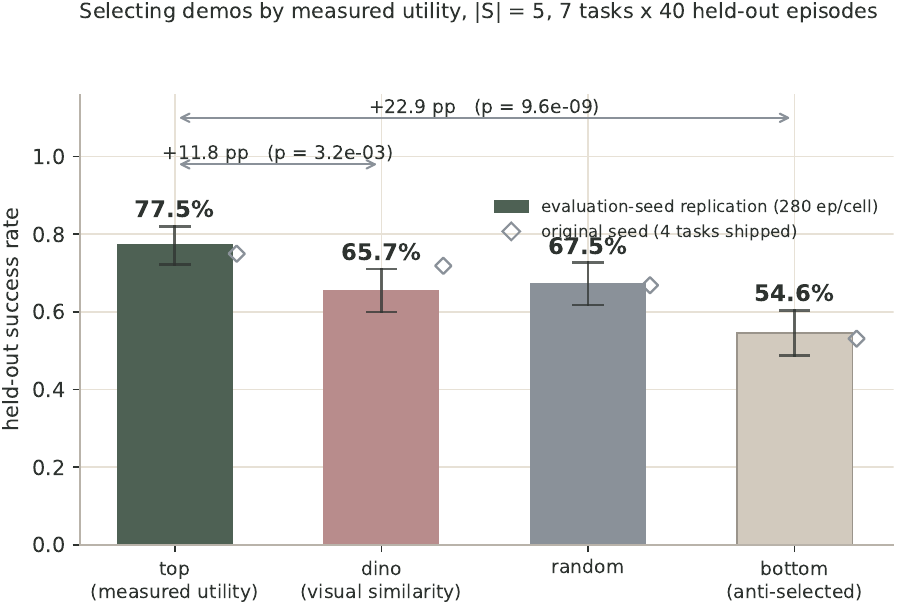}\end{minipage}\hfill
\begin{minipage}[t]{0.49\linewidth}\centering\includegraphics[width=\linewidth]{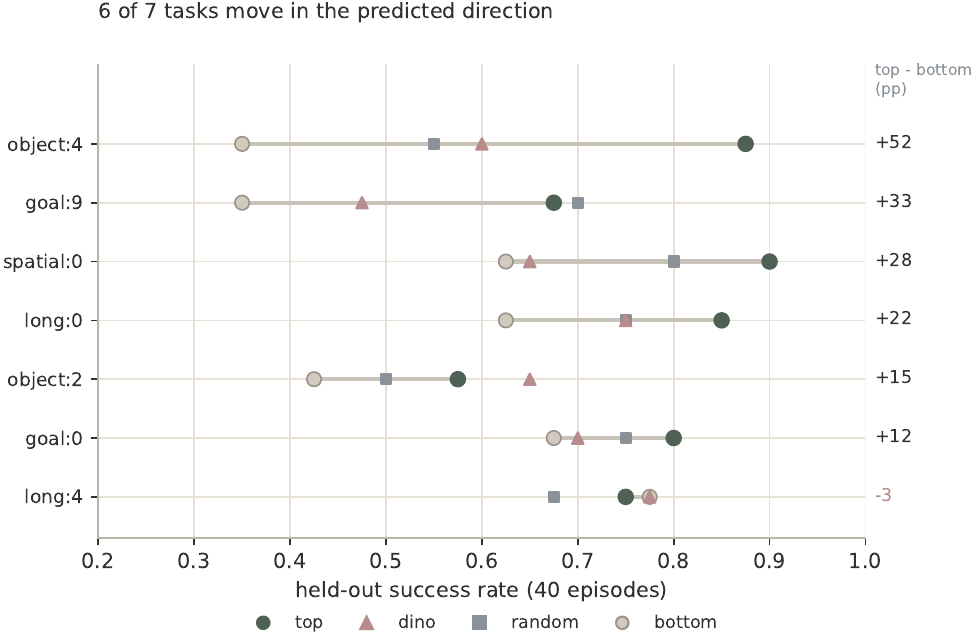}\end{minipage}
\caption{Seven-task family. Left: second evaluation seed (bars) with the four first-seed tasks whose episode files are recoverable (diamonds). Right: per-task cells; \ttt{long:4} is the documented outlier.}
\label{fig:seven}
\end{figure*}

\Cref{tab:pertask} gives every per-task cell behind the main text's \cref{tab:main,fig:main}, and \cref{tab:reliability} the split-half agreement of $\ph$ quoted in \cref{sec:why}. \Cref{fig:seven} shows the seven-task family across evaluation seeds; the context-necessity check of \cref{sec:gate} is \cref{fig:gate1} of the main text.

\begin{table*}[t]
\centering
\footnotesize
\setlength{\tabcolsep}{3.5pt}
\caption{Per-task held-out success (\%) at pool size 5, 40 episodes per cell, initial states 10--49, with the discordant counts $b$ (\ttt{top} wins) and $c$ (\ttt{dino} wins) for \ttt{top}$-$\ttt{dino}. Left block: the seven attribution-study tasks (second evaluation seed). Right block: the twelve unseen LIBERO-90 tasks.}
\label{tab:pertask}
\begin{tabular}{lcccccc|lcccccc}
\toprule
task & top & dino & random & bottom & $b$ & $c$ & task & top & dino & random & bottom & $b$ & $c$ \\
\midrule
spatial:0 & 90.0 & 65.0 & 80.0 & 62.5 & 12 & 2 & l90:0 & 77.5 & 67.5 & 72.5 & 55.0 & 7 & 3 \\
object:2 & 57.5 & 65.0 & 50.0 & 42.5 & 10 & 13 & l90:10 & 55.0 & 25.0 & 17.5 & 17.5 & 17 & 5 \\
object:4 & 87.5 & 60.0 & 55.0 & 35.0 & 14 & 3 & l90:12 & 42.5 & 35.0 & 35.0 & 10.0 & 11 & 8 \\
goal:0 & 80.0 & 70.0 & 75.0 & 67.5 & 8 & 4 & l90:15 & 42.5 & 35.0 & 42.5 & 25.0 & 11 & 8 \\
goal:9 & 67.5 & 47.5 & 70.0 & 35.0 & 15 & 7 & l90:20 & 32.5 & 20.0 & 22.5 & 15.0 & 9 & 4 \\
long:0 & 85.0 & 75.0 & 75.0 & 62.5 & 9 & 5 & l90:47 & 67.5 & 80.0 & 67.5 & 70.0 & 6 & 11 \\
long:4 & 75.0 & 77.5 & 67.5 & 77.5 & 8 & 9 & l90:55 & 60.0 & 47.5 & 47.5 & 45.0 & 7 & 2 \\
 & & & & & &  & l90:56 & 57.5 & 47.5 & 37.5 & 15.0 & 8 & 4 \\
 & & & & & &  & l90:59 & 47.5 & 40.0 & 50.0 & 27.5 & 10 & 7 \\
 & & & & & &  & l90:60 & 80.0 & 62.5 & 67.5 & 67.5 & 14 & 7 \\
 & & & & & &  & l90:61 & 67.5 & 67.5 & 52.5 & 37.5 & 8 & 8 \\
 & & & & & &  & l90:72 & 90.0 & 70.0 & 70.0 & 65.0 & 10 & 2 \\
\bottomrule
\end{tabular}
\end{table*}

\begin{table*}[t]
\centering
\footnotesize
\caption{Split-half Spearman agreement of $\ph$ (initial states 0--4 against 5--9, same pools) for the 20 attributed tasks, including the excluded \ttt{spatial:7}. Mean $0.139$; mean top-five Jaccard $0.20$.}
\label{tab:reliability}
\begin{tabular}{lc|lc|lc|lc}
\toprule
task & $\rho$ & task & $\rho$ & task & $\rho$ & task & $\rho$ \\
\midrule
spatial:0 & 0.663 & goal:9 & 0.268 & l90:0 & 0.316 & l90:47 & 0.367 \\
object:4 & 0.068 & spatial:7 & $-0.474$ & l90:55 & 0.002 & l90:59 & 0.083 \\
object:2 & 0.021 & long:0 & 0.383 & l90:72 & 0.329 & l90:12 & 0.535 \\
goal:0 & $-0.042$ & & & l90:20 & 0.107 & l90:10 & $-0.200$ \\
long:4 & 0.039 & & & l90:60 & 0.027 & l90:15 & 0.244 \\
 & & & & l90:56 & 0.065 & l90:61 & $-0.020$ \\
\bottomrule
\end{tabular}
\end{table*}

\paragraph{Repeat run against the shipped pool.}
\label{sup:repeat}
The three cells (\ttt{top}, \ttt{rerank}, \ttt{full}) were run again on the 7 study tasks and initial states 10--49 under a new evaluation seed (1115), with the existing $\ph$ and top-five lists (checked against the original fits) and no refitting, because the leads of \ttt{top} and \ttt{rerank} over \ttt{full} were not resolved at 280 paired episodes. The analysis was fixed before the run: an exact McNemar test on the 560 pairs pooling both runs, primary; the new run alone, secondary. The stack was rebuilt from the released scripts and checkpoint and passed a context-necessity check first (own pool 5/10, another task's pool 0/10 on \ttt{goal:0} states 0--9, as in the original campaign). New run: \ttt{full} 71.1\%, \ttt{rerank} 74.3\%, \ttt{top} 80.0\%; \ttt{top}$-$\ttt{full} $+8.9\pp$ ($b=61$, $c=36$, $p=0.014$), \ttt{rerank}$-$\ttt{full} $+3.2\pp$ ($b=45$, $c=36$, $p=0.37$). Pooled: $+7.0\pp$ ($b=117$, $c=78$, $p=0.006$) and $+4.5\pp$ ($b=91$, $c=66$, $p=0.055$). In the first run, \ttt{top} came from the second evaluation seed and \ttt{full} and \ttt{rerank} from the first, so its \ttt{top}$-$\ttt{full} pairs share task and initial state but not the environment seed; in the repeat all three share it.

\section{Additional figures}
\label{sup:figs}

Further figures cited from \cref{sec:prune,sec:why} of the main text.

\begin{figure*}[t]
\centering
\begin{minipage}[t]{0.55\linewidth}\centering\includegraphics[width=\linewidth]{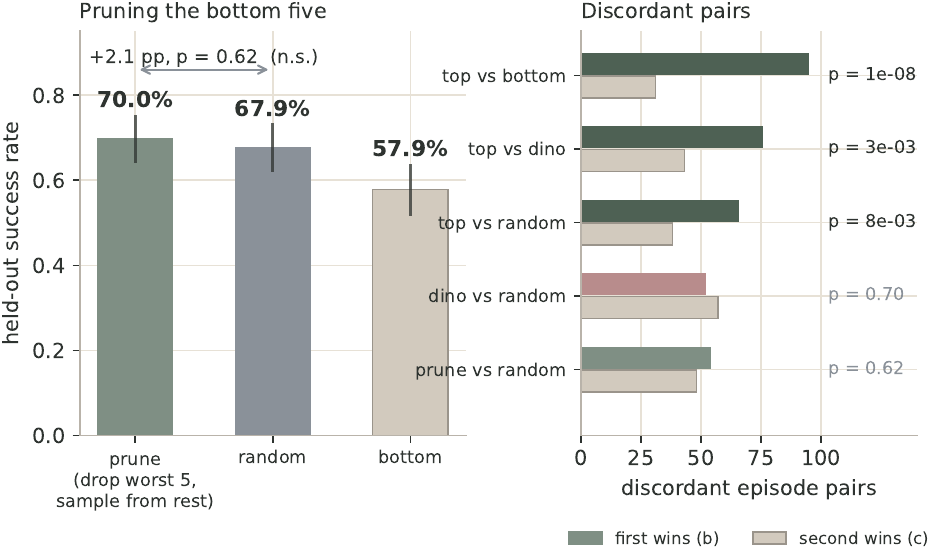}\end{minipage}\hfill
\begin{minipage}[t]{0.43\linewidth}\centering\includegraphics[width=\linewidth]{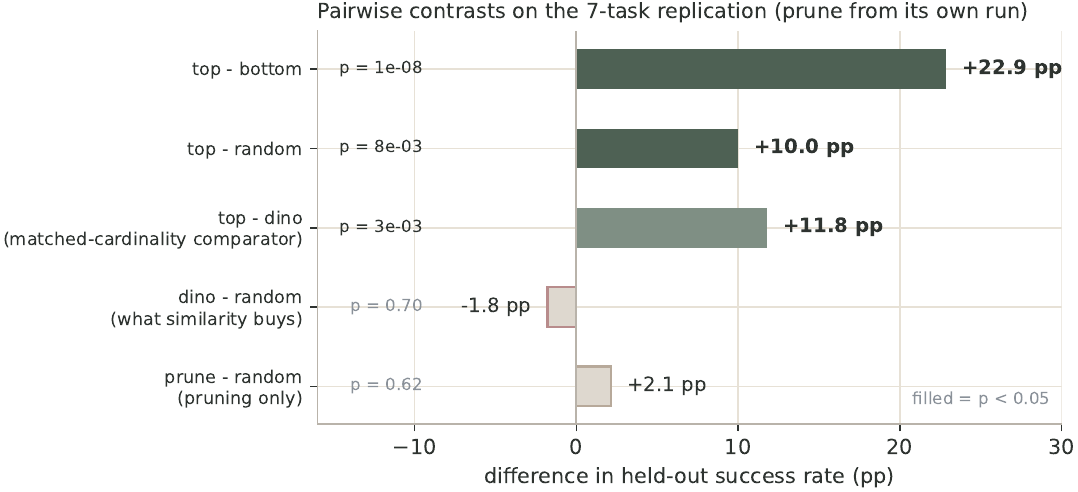}\end{minipage}
\caption{Left: the \ttt{prune} control against \ttt{random} and \ttt{bottom} from the same run, with the discordant-pair panel. Right: pairwise contrasts; the four size-5 cells come from the second-seed run and the \ttt{prune} contrast from its own run (hence random 67.9\% there against 67.5\%) (\cref{sec:prune}).}
\label{fig:prune}
\end{figure*}

\begin{figure*}[t]
\centering
\begin{minipage}[t]{0.49\linewidth}\centering\includegraphics[width=\linewidth]{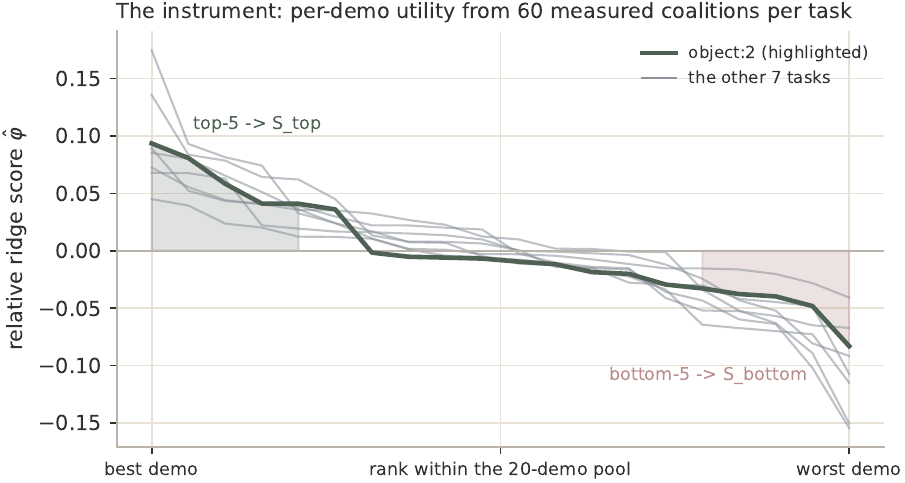}\end{minipage}\hfill
\begin{minipage}[t]{0.49\linewidth}\centering\includegraphics[width=\linewidth]{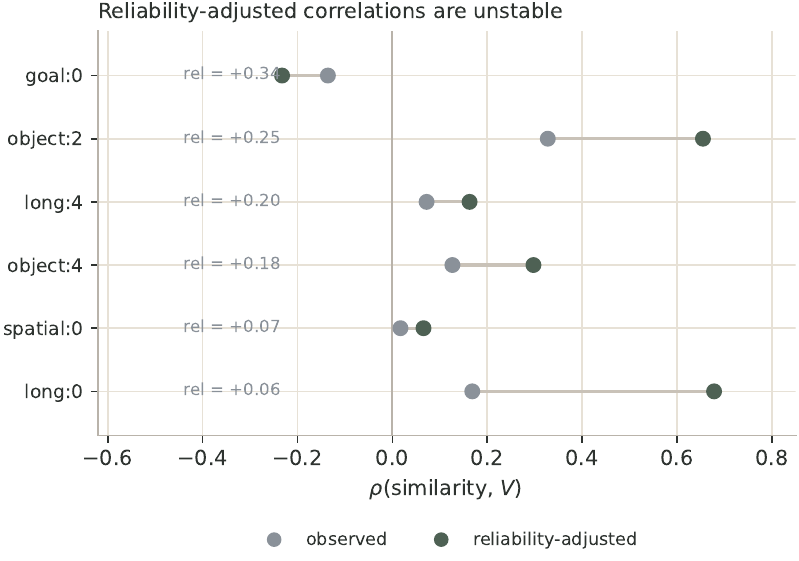}\end{minipage}
\caption{Left: sorted relative ridge scores $\ph$ per task; shaded ends form \ttt{top} and \ttt{bottom}. Right: split-half agreement of $\Vh$ per task and the reliability-adjusted similarity correlation, which is unstable.}
\label{fig:reliability}
\end{figure*}

\begin{figure*}[t]
\centering
\begin{minipage}[t]{0.49\linewidth}\centering\includegraphics[width=\linewidth]{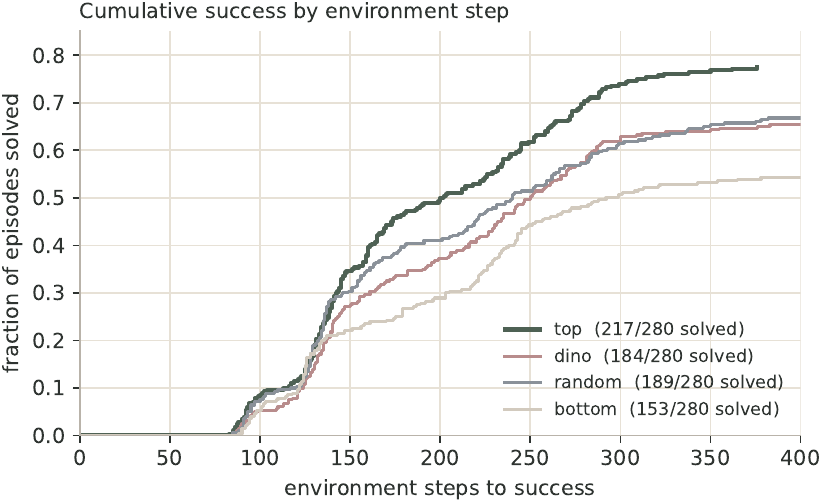}\end{minipage}\hfill
\begin{minipage}[t]{0.49\linewidth}\centering\includegraphics[width=\linewidth]{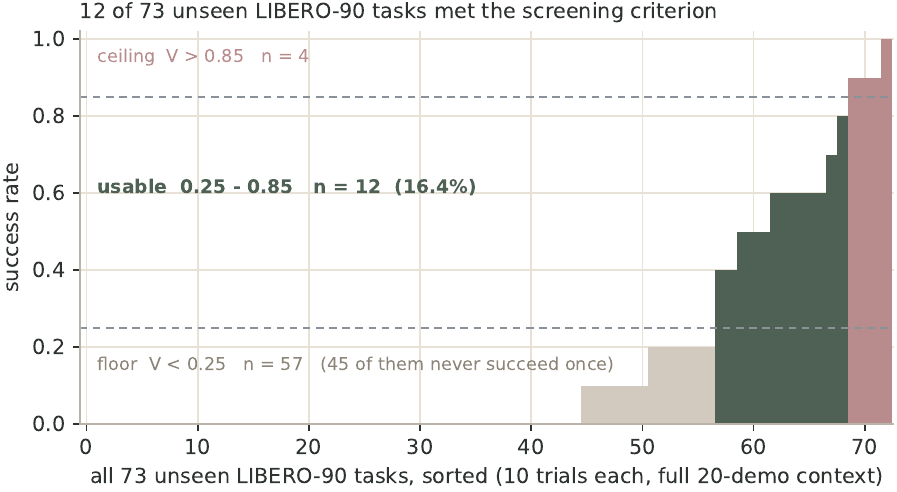}\end{minipage}
\caption{Left: cumulative success by step under \ttt{top} and \ttt{bottom}; the curves are similar below about 130 steps. Right: the 73 unseen LIBERO-90 tasks by full-pool success under the screen; 12 met the criterion.}
\label{fig:ecdf}
\end{figure*}

\section{Withdrawn hypotheses}
\label{sup:retracted}

\Cref{tab:retracted} of the main text lists the six readings that looked supported mid-study and were overturned by repeated measurement; \cref{fig:suite} of the main text shows the suite-structure case.

\section{Additional checkpoints}
\label{sup:ckpt2}

Three further published checkpoints of the same LIBERO port were run through the context-necessity check on four tasks prespecified at the median effect size of the 19-task study (\ttt{libero\_90:56}, \ttt{libero\_90:0}, \ttt{libero\_10:0}, \ttt{libero\_goal:0}), with the strongest task excluded. The LoRA-primed checkpoint at step 9\,000 (normalisation statistics repaired) succeeded on 1 of 40 episodes with matched context and 1 of 40 with mismatched; the LoRA checkpoint at step 4\,500 on 0 of 40 with matched context; the pre-priming in-context checkpoint at step 14\,999 on 4 of 40 in both arms, on the same four episodes. Our checkpoint under the same protocol reaches 60\% on the two LIBERO-90 tasks. None of the three discriminates matched from mismatched context on these tasks, so no selection cell was run; dead adapters and load failures were ruled out. On tasks chosen for our checkpoint's eligibility band, the pre-priming checkpoint's 4 successes in 40 fit both a policy that ignores its context and one that cannot do these tasks.

\section{Retrieval-frequency associations (exploratory)}
\label{sup:traces}

Over 11\,200 (demonstration, episode) rows from the two logged retrieval modes, regressing per-episode success on a demonstration's retrieval share, its $\ph$ and their interaction gives coefficients of $-0.09$ (share), $-0.006$ ($\ph$) and $+2.16$ (interaction); the correlation between retrieval share and $\ph$ is $+0.28$. Rows repeat episode outcomes across demonstrations and retrieval share depends on the evolving trajectory, so the coefficients describe associations.

\section{Reproducibility statement}
\label{sup:repro}

The campaign scripts (value function, attribution, selection cells, similarity analyses, reliability and confound checks, LIBERO-90 screen, the four server patches and their application order) and the per-episode records behind every cell of the second-seed 7-task family, the 12-task extension, the size-3 and size-10 sweeps, the prune, full-pool and budget analyses, and the repeat run of \cref{sup:repeat} are provided as supplementary material; the first evaluation seed is available in aggregate only because three of its per-task episode files are lost. The seven-task study used 170 GPU-hours, 66.9 of them on attribution; the repeat run's 840 episodes took 3.4 hours on one GPU. Each released cell recomputes from raw episodes; rollouts are cached by (checkpoint, task, sorted pool, initial-state set, step cap). The policy, its LIBERO port and the preprocessed pools are public. Rendering requires OSMesa on hosts without an NVIDIA EGL driver; the DINOv2 hub loader must be patched to local mode because it contacts GitHub even when weights are cached.

\end{document}

%% file: preamble.tex
\usepackage{cuted} 

\usepackage{currfile} 

\usepackage{caption} 



%% file: math_commands.tex
\usepackage{amsmath,amsfonts,bm}

\def\eqref#1{equation~\ref{#1}}

\def\1{\bm{1}}

\DeclareMathAlphabet{\mathsfit}{\encodingdefault}{\sfdefault}{m}{sl}
\SetMathAlphabet{\mathsfit}{bold}{\encodingdefault}{\sfdefault}{bx}{n}



%% file: references.bib
@article{sridhar2025ricl,
  title={Ricl: Adding in-context adaptability to pre-trained vision-language-action models},
  author={Sridhar, Kaustubh and Dutta, Souradeep and Jayaraman, Dinesh and Lee, Insup},
  journal={arXiv preprint arXiv:2508.02062},
  year={2025}
}

@article{pertsch2025fast,
  title={Fast: Efficient action tokenization for vision-language-action models},
  author={Pertsch, Karl and Stachowicz, Kyle and Ichter, Brian and Driess, Danny and Nair, Suraj and Vuong, Quan and Mees, Oier and Finn, Chelsea and Levine, Sergey},
  journal={arXiv preprint arXiv:2501.09747},
  year={2025}
}

@article{black2024pi_0,
  title={$\pi_0$: A Vision-Language-Action Flow Model for General Robot Control},
  author={Black, Kevin and Brown, Noah and Driess, Danny and Esmail, Adnan and Equi, Michael and Finn, Chelsea and Fusai, Niccolo and Groom, Lachy and Hausman, Karol and Ichter, Brian and others},
  journal={arXiv preprint arXiv:2410.24164},
  year={2024}
}

@article{liu2023libero,
  title={Libero: Benchmarking knowledge transfer for lifelong robot learning},
  author={Liu, Bo and Zhu, Yifeng and Gao, Chongkai and Feng, Yihao and Liu, Qiang and Zhu, Yuke and Stone, Peter},
  journal={Advances in Neural Information Processing Systems},
  volume={36},
  pages={44776--44791},
  year={2023}
}

@article{oquab2023dinov2,
  title={Dinov2: Learning robust visual features without supervision},
  author={Oquab, Maxime and Darcet, Timoth{\'e}e and Moutakanni, Th{\'e}o and Vo, Huy and Szafraniec, Marc and Khalidov, Vasil and Fernandez, Pierre and Haziza, Daniel and Massa, Francisco and El-Nouby, Alaaeldin and others},
  journal={arXiv preprint arXiv:2304.07193},
  year={2023}
}

@article{douze2025faiss,
  title={The faiss library},
  author={Douze, Matthijs and Guzhva, Alexandr and Deng, Chengqi and Johnson, Jeff and Szilvasy, Gergely and Mazar{\'e}, Pierre-Emmanuel and Lomeli, Maria and Hosseini, Lucas and J{\'e}gou, Herv{\'e}},
  journal={IEEE Transactions on Big Data},
  year={2025},
  publisher={IEEE}
}

@inproceedings{vosylius2025instant,
  title={Instant policy: In-context imitation learning via graph diffusion},
  author={Vosylius, Vitalis and Johns, Edward},
  booktitle={International Conference on Learning Representations},
  volume={2025},
  pages={35011--35030},
  year={2025}
}

@article{fu2024context,
  title={In-context imitation learning via next-token prediction},
  author={Fu, Letian and Huang, Huang and Datta, Gaurav and Chen, Lawrence Yunliang and Panitch, William Chung-Ho and Liu, Fangchen and Li, Hui and Goldberg, Ken},
  journal={arXiv preprint arXiv:2408.15980},
  year={2024}
}

@article{di2024keypoint,
  title={Keypoint action tokens enable in-context imitation learning in robotics},
  author={Di Palo, Norman and Johns, Edward},
  journal={arXiv preprint arXiv:2403.19578},
  year={2024}
}

@inproceedings{yin2025context,
  title={In-context learning enables robot action prediction in llms},
  author={Yin, Yida and Wang, Zekai and Sharma, Yuvan and Niu, Dantong and Darrell, Trevor and Herzig, Roei},
  booktitle={2025 IEEE International Conference on Robotics and Automation (ICRA)},
  pages={8972--8979},
  year={2025},
  organization={IEEE}
}

@inproceedings{xu2022prompting,
  title={Prompting decision transformer for few-shot policy generalization},
  author={Xu, Mengdi and Shen, Yikang and Zhang, Shun and Lu, Yuchen and Zhao, Ding and Tenenbaum, Joshua and Gan, Chuang},
  booktitle={international conference on machine learning},
  pages={24631--24645},
  year={2022},
  organization={PMLR}
}

@article{mandlekar2021matters,
  title={What matters in learning from offline human demonstrations for robot manipulation},
  author={Mandlekar, Ajay and Xu, Danfei and Wong, Josiah and Nasiriany, Soroush and Wang, Chen and Kulkarni, Rohun and Fei-Fei, Li and Savarese, Silvio and Zhu, Yuke and Mart{\'\i}n-Mart{\'\i}n, Roberto},
  journal={arXiv preprint arXiv:2108.03298},
  year={2021}
}

@article{du2023behavior,
  title={Behavior retrieval: Few-shot imitation learning by querying unlabeled datasets},
  author={Du, Maximilian and Nair, Suraj and Sadigh, Dorsa and Finn, Chelsea},
  journal={arXiv preprint arXiv:2304.08742},
  year={2023}
}

@inproceedings{memmel2025strap,
  title={Strap: Robot sub-trajectory retrieval for augmented policy learning},
  author={Memmel, Marius and Berg, Jacob and Chen, Bingqing and Gupta, Abhishek and Francis, Jonathan},
  booktitle={International Conference on Learning Representations},
  volume={2025},
  pages={63813--63832},
  year={2025}
}

@article{lin2024flowretrieval,
  title={Flowretrieval: Flow-guided data retrieval for few-shot imitation learning},
  author={Lin, Li-Heng and Cui, Yuchen and Xie, Amber and Hua, Tianyu and Sadigh, Dorsa},
  journal={arXiv preprint arXiv:2408.16944},
  year={2024}
}

@article{hejna2024re,
  title={Re-mix: Optimizing data mixtures for large scale imitation learning},
  author={Hejna, Joey and Bhateja, Chethan and Jiang, Yichen and Pertsch, Karl and Sadigh, Dorsa},
  journal={arXiv preprint arXiv:2408.14037},
  year={2024}
}

@article{nguyen2023context,
  title={In-context example selection with influences},
  author={Nguyen, Tai and Wong, Eric},
  journal={arXiv preprint arXiv:2302.11042},
  year={2023}
}

@inproceedings{xie2025demoshapley,
  title={Demoshapley: Valuation of demonstrations for in-context learning},
  author={Xie, Shan and Luo, Man and Stern, Chadly Daniel and Du, Mengnan and Lu, Cheng},
  booktitle={2025 IEEE International Conference on Big Data (BigData)},
  pages={4081--4090},
  year={2025},
  organization={IEEE}
}

@article{belkhale2023data,
  title={Data quality in imitation learning},
  author={Belkhale, Suneel and Cui, Yuchen and Sadigh, Dorsa},
  journal={Advances in neural information processing systems},
  volume={36},
  pages={80375--80395},
  year={2023}
}

@inproceedings{lin2025data,
  title={Data scaling laws in imitation learning for robotic manipulation},
  author={Lin, Fanqi and Hu, Yingdong and Sheng, Pingyue and Wen, Chuan and You, Jiacheng and Gao, Yang},
  booktitle={International Conference on Learning Representations},
  volume={2025},
  pages={54877--54910},
  year={2025}
}

@inproceedings{liu2022makes,
  title={What makes good in-context examples for GPT-3?},
  author={Liu, Jiachang and Shen, Dinghan and Zhang, Yizhe and Dolan, William B and Carin, Lawrence and Chen, Weizhu},
  booktitle={Proceedings of Deep Learning Inside Out (DeeLIO 2022): The 3rd workshop on knowledge extraction and integration for deep learning architectures},
  pages={100--114},
  year={2022}
}

@inproceedings{rubin2022learning,
  title={Learning to retrieve prompts for in-context learning},
  author={Rubin, Ohad and Herzig, Jonathan and Berant, Jonathan},
  booktitle={Proceedings of the 2022 conference of the North American chapter of the association for computational linguistics: human language technologies},
  pages={2655--2671},
  year={2022}
}

@inproceedings{lu2022fantastically,
  title={Fantastically ordered prompts and where to find them: Overcoming few-shot prompt order sensitivity},
  author={Lu, Yao and Bartolo, Max and Moore, Alastair and Riedel, Sebastian and Stenetorp, Pontus},
  booktitle={Proceedings of the 60th Annual Meeting of the Association for Computational Linguistics (Volume 1: Long Papers)},
  pages={8086--8098},
  year={2022}
}

@inproceedings{workrethinking,
  title={Rethinking the Role of Demonstrations: What Makes In-Context Learning Work?},
  author={Min, Sewon and Lyu, Xinxi and Holtzman, Ari and Artetxe, Mikel and Lewis, Mike and Hajishirzi, Hannaneh and Zettlemoyer, Luke},
  booktitle={Proceedings of the 2022 Conference on Empirical Methods in Natural Language Processing},
  year={2022}
}

@inproceedings{ye2023compositional,
  title={Compositional exemplars for in-context learning},
  author={Ye, Jiacheng and Wu, Zhiyong and Feng, Jiangtao and Yu, Tao and Kong, Lingpeng},
  booktitle={International Conference on Machine Learning},
  pages={39818--39833},
  year={2023},
  organization={PMLR}
}

@inproceedings{zhang2022active,
  title={Active example selection for in-context learning},
  author={Zhang, Yiming and Feng, Shi and Tan, Chenhao},
  booktitle={Proceedings of the 2022 Conference on Empirical Methods in Natural Language Processing},
  pages={9134--9148},
  year={2022}
}

@inproceedings{dong2024survey,
  title={A survey on in-context learning},
  author={Dong, Qingxiu and Li, Lei and Dai, Damai and Zheng, Ce and Ma, Jingyuan and Li, Rui and Xia, Heming and Xu, Jingjing and Wu, Zhiyong and Chang, Baobao and others},
  booktitle={Proceedings of the 2024 conference on empirical methods in natural language processing},
  pages={1107--1128},
  year={2024}
}

@inproceedings{ghorbani2019data,
  title={Data shapley: Equitable valuation of data for machine learning},
  author={Ghorbani, Amirata and Zou, James},
  booktitle={International conference on machine learning},
  pages={2242--2251},
  year={2019},
  organization={PMLR}
}

@inproceedings{jia2019towards,
  title={Towards efficient data valuation based on the shapley value},
  author={Jia, Ruoxi and Dao, David and Wang, Boxin and Hubis, Frances Ann and Hynes, Nick and G{\"u}rel, Nezihe Merve and Li, Bo and Zhang, Ce and Song, Dawn and Spanos, Costas J},
  booktitle={The 22nd international conference on artificial intelligence and statistics},
  pages={1167--1176},
  year={2019},
  organization={PMLR}
}

@article{kwon2021beta,
  title={Beta shapley: a unified and noise-reduced data valuation framework for machine learning},
  author={Kwon, Yongchan and Zou, James},
  journal={arXiv preprint arXiv:2110.14049},
  year={2021}
}

@inproceedings{wang2023data,
  title={Data banzhaf: A robust data valuation framework for machine learning},
  author={Wang, Jiachen T and Jia, Ruoxi},
  booktitle={International conference on artificial intelligence and statistics},
  pages={6388--6421},
  year={2023},
  organization={PMLR}
}

@inproceedings{koh2017understanding,
  title={Understanding black-box predictions via influence functions},
  author={Koh, Pang Wei and Liang, Percy},
  booktitle={International conference on machine learning},
  pages={1885--1894},
  year={2017},
  organization={Pmlr}
}

@article{pruthi2020estimating,
  title={Estimating training data influence by tracing gradient descent},
  author={Pruthi, Garima and Liu, Frederick and Kale, Satyen and Sundararajan, Mukund},
  journal={Advances in neural information processing systems},
  volume={33},
  pages={19920--19930},
  year={2020}
}

@article{ilyas2022datamodels,
  title={Datamodels: Predicting predictions from training data},
  author={Ilyas, Andrew and Park, Sung Min and Engstrom, Logan and Leclerc, Guillaume and Madry, Aleksander},
  journal={arXiv preprint arXiv:2202.00622},
  year={2022}
}

@article{park2023trak,
  title={Trak: Attributing model behavior at scale},
  author={Park, Sung Min and Georgiev, Kristian and Ilyas, Andrew and Leclerc, Guillaume and Madry, Aleksander},
  journal={arXiv preprint arXiv:2303.14186},
  year={2023}
}

@article{johnson2019billion,
  title={Billion-scale similarity search with GPUs},
  author={Johnson, Jeff and Douze, Matthijs and J{\'e}gou, Herv{\'e}},
  journal={IEEE transactions on big data},
  volume={7},
  number={3},
  pages={535--547},
  year={2019},
  publisher={IEEE}
}

@article{mcnemar1947note,
  title={Note on the sampling error of the difference between correlated proportions or percentages},
  author={McNemar, Quinn},
  journal={Psychometrika},
  volume={12},
  number={2},
  pages={153--157},
  year={1947},
  publisher={Cambridge University Press \& Assessment}
}

@article{cochran1954combination,
  title={The combination of estimates from different experiments},
  author={Cochran, William G},
  journal={Biometrics},
  volume={10},
  number={1},
  pages={101--129},
  year={1954},
  publisher={JSTOR}
}

@article{higgins2002quantifying,
  title={Quantifying heterogeneity in a meta-analysis},
  author={Higgins, Julian PT and Thompson, Simon G},
  journal={Statistics in medicine},
  volume={21},
  number={11},
  pages={1539--1558},
  year={2002},
  publisher={Wiley Online Library}
}

@article{wilson1927probable,
  title={Probable inference, the law of succession, and statistical inference},
  author={Wilson, Edwin B},
  journal={Journal of the American Statistical Association},
  volume={22},
  number={158},
  pages={209--212},
  year={1927},
  publisher={Taylor \& Francis}
}

@inproceedings{chang2023data,
  title={Data curation alone can stabilize in-context learning},
  author={Chang, Ting-Yun and Jia, Robin},
  booktitle={Proceedings of the 61st Annual Meeting of the Association for Computational Linguistics (Volume 1: Long Papers)},
  pages={8123--8144},
  year={2023}
}

@inproceedings{sridhar2025regent,
  title={REGENT: A retrieval-augmented generalist agent that can act in-context in new environments},
  author={Sridhar, Kaustubh and Dutta, Souradeep and Jayaraman, Dinesh and Lee, Insup},
  booktitle={International Conference on Learning Representations},
  volume={2025},
  pages={39385--39414},
  year={2025}
}

@inproceedings{wang2023images,
  title={Images speak in images: A generalist painter for in-context visual learning},
  author={Wang, Xinlong and Wang, Wen and Cao, Yue and Shen, Chunhua and Huang, Tiejun},
  booktitle={2023 IEEE/CVF Conference on Computer Vision and Pattern Recognition (CVPR)},
  pages={6830--6839},
  year={2023},
  organization={IEEE}
}

@article{wang2023seggpt,
  title={Seggpt: Segmenting everything in context},
  author={Wang, Xinlong and Zhang, Xiaosong and Cao, Yue and Wang, Wen and Shen, Chunhua and Huang, Tiejun},
  journal={arXiv preprint arXiv:2304.03284},
  year={2023}
}

@article{zhang2023makes,
  title={What makes good examples for visual in-context learning?},
  author={Zhang, Yuanhan and Zhou, Kaiyang and Liu, Ziwei},
  journal={Advances in Neural Information Processing Systems},
  volume={36},
  pages={17773--17794},
  year={2023}
}

@article{sun2025exploring,
  title={Exploring effective factors for improving visual in-context learning},
  author={Sun, Yanpeng and Chen, Qiang and Wang, Jian and Wang, Jingdong and Li, Zechao},
  journal={IEEE Transactions on Image Processing},
  year={2025},
  publisher={IEEE}
}

@article{xu2026athena,
  title={ATHENA: Accelerated Multi-Task Heterogeneous Influence Functions for Robot Data Curation},
  author={Xu, Tao and Wang, Jiaxin and Zhang, Runhao and Guan, Jiayi and Zeng, Xianchao and Song, Weixi and Zhou, Xinyu and Chen, Zhetao and Chen, Guang and Li, Yong-Lu},
  journal={arXiv preprint arXiv:2606.16208},
  year={2026}
}

@article{jiang2026we,
  title={What Are We Actually Benchmarking in Robot Manipulation?},
  author={Jiang, Tianchong and Tan, Xiangshan and Wheeler, Samuel and Sun, Luzhe and Ayalew, Tewodros W and Walter, Matthew},
  journal={arXiv preprint arXiv:2606.04233},
  year={2026}
}
